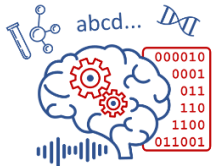

*Research article*

# Crop and weed classification based on AutoML

**Xuetao Jiang[1], Binbin Yong[1], Soheila Garshasbi[2], Jun Shen[2], Meiyu Jiang[1] and Qingguo Zhou[1,*]**

[1] School of Information Science and Engineering, Lanzhou University, Lanzhou, 730000, China
[2] School of Computing and Information Technology, University of Wollongong, NSW 2522, Australia

**\* Correspondence:** zhouqg@lzu.edu.cn

Academic editor: Chih-Cheng Hung

**Abstract:** CNN models already play an important role in classification of crop and weed with high accuracy, more than 95% as reported in literature. However, to manually choose and fine-tune the deep learning models becomes laborious and indispensable in most traditional practices and research. Moreover, the classic objective functions are not thoroughly compatible with agricultural farming tasks as the corresponding models suffer from misclassifying crop to weed, often more likely than in other deep learning application domains. In this paper, we applied autonomous machine learning with a new objective function for crop and weed classification, achieving higher accuracy and lower crop killing rate (rate of identifying a crop as a weed). The experimental results show that our method outperforms state-of-the-art applications, for example, ResNet and VGG19.

## 1. Introduction

Weeding has been a big issue faced by farmers for a long time, especially for those who work in gigantic farms. Industrialized modern agriculture applies chemical methods to control weed; however, it leads to a great increase in herbicide resistance and rising harms to the ecological environment. Therefore, more focused research is required on smart and intelligent technologies for precision weed management. Meanwhile, effective farming practices also demand precision crop cultivation to increase the overall agricultural yields. Under both the conditions mentioned herein, each kind of plant should be treated with different strategies. As it is time consuming and laborious to conduct this process manually, a great deal of effort has been put into facilitating autonomous farming in which various computational intelligence models are designed to classify field plants.

Implementation of the autonomous farming has been a research focus in modern agriculture, especially for the emerging Agriculture-4.0 [1]; however, plant classification is still a challenging research issue in this realm [2,3]. Many studies on weed and crop classification have been carried out over the past few decades to meet the needs for precision weed management. Generally, these studies can be divided into three categories: 3D point cloud classification, spectrum classification and image classification. 3D point cloud classification relies on intensive computing to determine the label of each entity, which requires 3D Li-DAR data [4] with their bounding boxes. For the spectrum classification, monochrome cameras with different lasers were used to match the spectral reflectance for classification [5]. In image classification [6], cameras were used to capture images of the field. 3D features and spectrum features could be appropriate factors for weed classification, but high-priced instrumentation and computing devices would be required, not suitable for every farmer.

Utilizing images for crop and weed classification can be implemented by traditional algorithm and convolution neural network (CNN). In many image classification tasks, CNN has higher accuracy than traditional algorithm like KNN, SVM and MLP [7], which also produce remarkable results in crop and weed classification. However, there is no published study focusing on applying auto machine learning (AutoML) [8] on crop and weed classification, nor on training models with high accuracy and low crop killing rate (CKR, rate of identifying a crop as a weed).

This paper presents a method of crop and weed classification based on AutoML and ensemble modeling, in order to achieve better performance on the outdoor greenhouse data set without manual model selection. Compared to existing methods, it consists of four improvements:

- Using AutoML to select the optimal model automatically;
- New metric in evaluation and new objective function in model training;
- New algorithm for the compatible model among different data sets;
- Ensemble strategy to reaching high accuracy and low *CKR*.

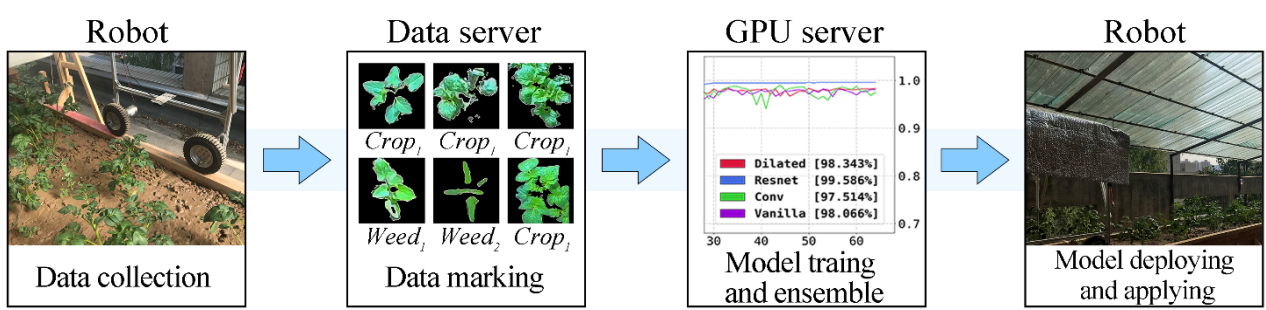

**Figure 1.** Pipeline of crop and weed classification.

The pipeline of this project is shown in Fig. 1. First, image data are collected and preprocessed by a robot in greenhouse. Then, the processed images are uploaded to a data server, and labeled manually. The labeled data is sent to GPU server for models training. Finally, the trained models are gathered and deployed on the robot to perform detection tasks. The whole system is automatic except for data labeling.

## 2. Materials and methods

As presented in Fig. 2, our method includes four steps: data acquisition, image pre-processing, model searching and training, ensemble modeling. We build a new data set in two steps: data acquisition and image pre-processing. For model searching and training, we apply AutoML based algorithm and new objective function. At the end, we use ensemble model to make final prediction.

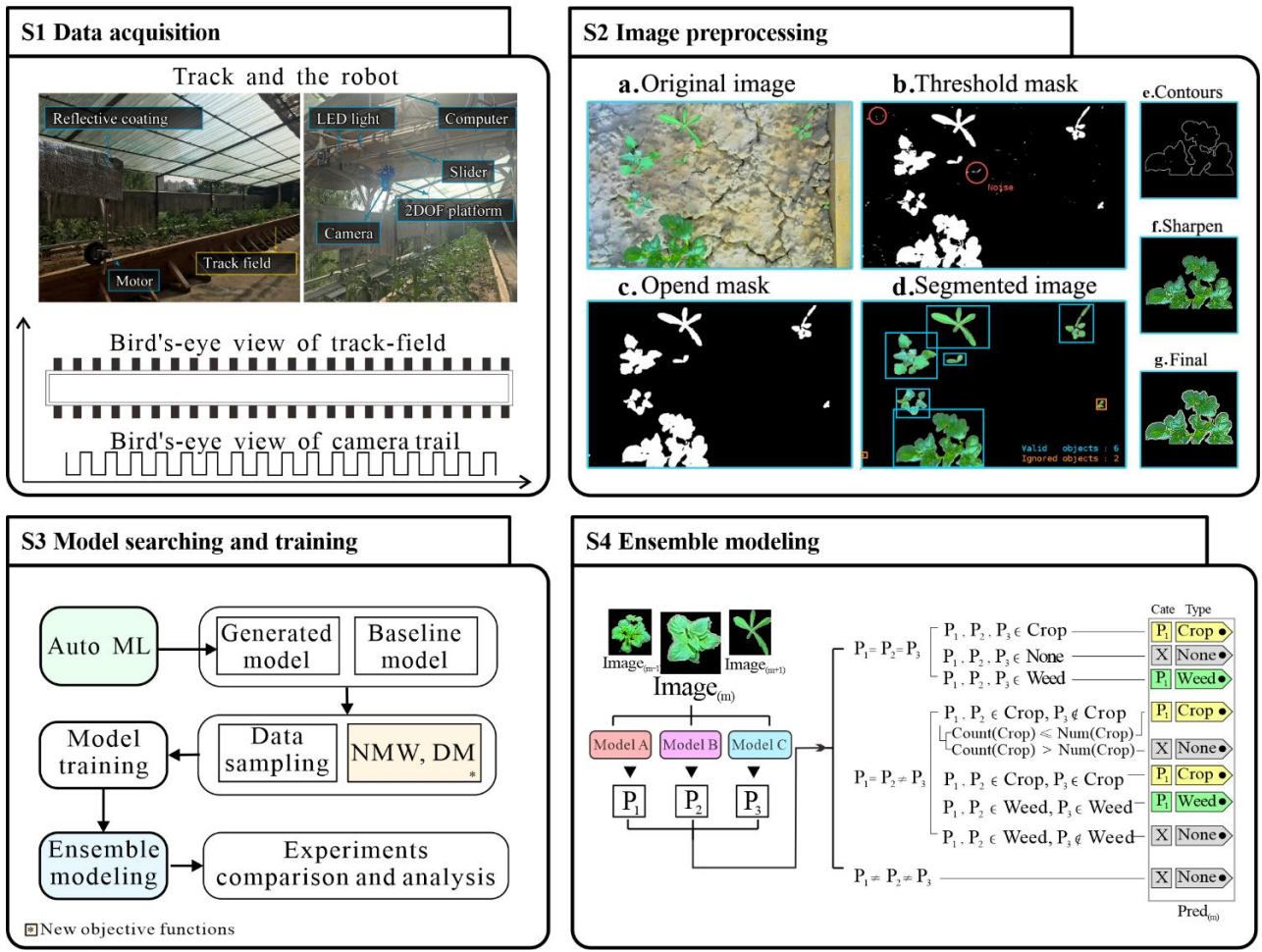

**Figure 2.** The experiment framework and flow chart.

### 2.1. Data acquisition and Image pre-processing

In this work, our first step is to construct the data set. Since there is no farmland near our university campus, we built a wooden cuboid box with board, screws, waterproof membrane and used it to plant potatoes and two kinds of weeds, which is shown in Fig. 2 S1. A four wheeled robot is built to collect data, and its hardware system includes raspberry Pi, USB camera, motor and control chip. The robot can traverse in the field step by step, while the slider moves the camera horizontally. Equipped with a kit of LED lights, the robot can also perform satisfactorily on an overcast day. It collects raw data from the field, and subsequently we pre-process the data at the next step.

We use Algorithm 1 to extract crops and weeds images from field images, then we label them manually to construct the data set. In Algorithm 1, functions *rgb2hsv, mask, morphologyOpen* and *findCounters* are based on OpenCV [9], and the results of key procedures are shown in Fig. 2 S2.

---

**Algorithm 1** Field image segmentation

---

**Input**: img, $T_{size}$, $T_{ratio}$
**Output**: Seg={Seg1, Seg2, · · ·, Segm};
1: hsv_img = rgb2hsv(img)
2: masked_img = mask(hsv_img)
3: opened_img = morphologyOpen(masked_img)
4: counters = findCounters(opened_img)
5: Seg = []
6: **for** counter ∈ counters **do**
7:    obj_img = boundingBoxCut (img, counter)
8:    **if** size(obj_img) / size(img) < $T_{size}$ **then**
9:        **continue**
10:   **end if**
11:   **if** ratio(obj_img) < $T_{ratio}$ **then**
12:       **continue**
13:   **end if**
14:   sharpen_img = sharpen(obj_img)
15:   out_img = bitwiseAnd(sharpen_img, counter)
16:   Seg.append(out_img)
17: **end for**
18: **return** Seg

---

As presented in Algorithm 1, the algorithm takes original field image *img*, threshold of image size as $T_{size}$ , threshold of image shape's ratio as $T_{ratio}$ (e.g., for an image of shape (200,100,1) or (100,200,1), its ratio is 0.5 ). From line 1 to 2, we convert RGB to HSV and mask pixels with Equation (1) listed below, which removes the background of image as S2.b of Fig. 2. In line 3, we use morphology open method to remove noise in mask as S2.c of Fig. 2. From line 8 to 13, we separate the crops and weeds in images and skip the small/banded objects as S2.d of Fig. 2. From line 14 to 15, we highlight the counter of sharpen images as S2.g of Fig. 2. Then we label these images manually to establish our data set.

$$mask\big(Pixel_{(i,j)}\big) = Pixel_{(i,j)} \begin{cases} H_{(i,j)} \in (45,95] \\ S_{(i,j)} \in (55,255] \\ V_{(i,j)} \in (55,255] \end{cases} \tag{1}$$

For Equation (1), we cluster images according to luminous intensity provided by the light sensor, then analyze the histogram of each cluster to determine the range of HSV.

## 2.2. Model searching and training

The flow chart of model searching and training is shown in Fig. 2 S3, and the key parts are marked with different colors, which will be detailed in the next three subsections.

The model search process employs AutoML-based algorithm to generate several models on two data sets, and we apply new objective functions on models training. In the first subsection, we will introduce two data sets used in this paper and the sampling rule. Then, we elaborate the AutoML-based algorithm in the second subsection. For the model training section, we detail serval objective functions and apply them on models training.

### 2.2.1. Data set detail

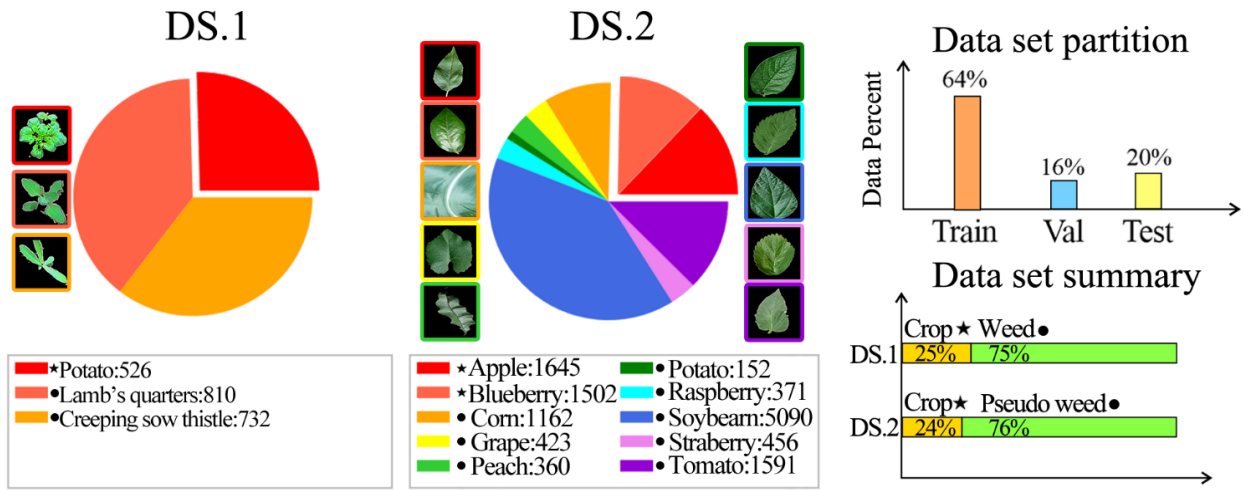

**Figure 3.** Plants percentage and data set partition.

The proposed methodology of crop and weed classification is carried out on two data sets: *DS.1*, a new data set collected in a small field by an agricultural robot, with 2068 images of potatoes and weeds, and in addition to *DS.2*, an open source big data set of crops called PlantVillage2 [10], with12,752 images of different crops, was also used. Although there are no weeds in *DS.2*, we can manually choose some crops as weeds, given that they do not affect the performance of models.

With a motivation to simulate an environment of non-chemical weeding, we do not apply weed control during the data collection, i.e. there is no plastic mulch or herbicide, which might lead to more weed than crop in biomass. After image pre-processing and manual labeling, we capture data set *DS*.1 with weed to crop ratio of approximately 2:1, as indicated in Fig. 3. An imbalanced data set affects the prediction of models, as a model tends to be misled by classifying an object as a weed to get higher score. Therefore, we use sampling approaches to produce the weed sub sets from the original complete data set (CD). For each class of weed in the data set, we calculate sample rate *k* as Equation (2), and concreate sampled weed part and crop part to get a sampled data set (SD) with a ratio of weed to crop approximately 1:1. In this paper, both *DS.1* and *DS.2* are applied with CD and SD to test the feasibility and robustness of our methodology.

$$k = \alpha \cdot \frac{Num_{\text{weed}\_i}}{Num_{weed}} + \beta \cdot \frac{1}{Cls_{weed}} \qquad (2)$$

where $Num_{weed}$ is the volume of weed data, $Num_{weed\_i}$ is the volume of specific weed, $Cls_{weed}$ is the classes of weed, $\alpha$ and $\beta$ are parameters set to 0.7 and 0.3.

### 2.2.2. Searching an optimal model

We use AutoKeras [11] as the AutoML framework for autonomous computing instead of testing models manually. AutoKeras is an open source package widely used for image and text classification. Deep models are feasible on powerful GPU in laboratory [12], but not on single board computer in farm robot. Hence, we restricted the model in AutoML to 300,000 parameters.

Given a type of models, AutoML might yield totally different results for different data set, that would cause inconsistency in the next step. We hence proposed Algorithm 2 to make the searching procedure feasible in multiple data sets, which allowed us to get similar model structure besides the output layer. For models of the same type, there is only a slight structure divergence in the last two layers, as different data sets have unequal classes.

---

**Algorithm 2** Optimal models in multiple data set

---

**Input**: T={T$_1$, T$_2$, • • • ,T$_m$}, DS={DS.1, DS.2, · · ; DS.n}
**Output**: M={M1, M2, · · ·,Mm};
1: **M** = ∅
2: **for** t ∈ T **do**
3:     **for** ds ∈ *DS* **do**
4:         trails$_t^{ds}$ = search (t, ds)
5:     **end for**
6:     Common = trails$_t^{DS.1}$ ∩ trails$_t^{DS.2}$ ∩ · · · ∩ trails$_t^{DS.n}$
7:     **if** Common ≠ ∅ **then**
8:         best$_t$ = max(common)
9:     **else**
10:        score_table = ∅
11:        **for** ds ∈ *DS* **do**
12:           score_table += evaluate(trails$_t^{ds}$ ,*DS*)
13:        **end for**
14:        best$_t$ = max(score_table)
15:     **end if**
16:     **M**$_t$ = best$_t$
17: **end for**
18: **return** M

---

As shown in Algorithm 2, the algorithm takes AutoML model type list *T*, data set list *DS* as the input. It outputs a list of optimal models *M* in the given data sets and model types. In line 4, *search* function is used for searching models, which requires type of model *t*, current data set *ds*. Return of *search* function is an optimal CNN model derived from AutoKeras. In line 6, the operator of set intersection ∩ is used to get common part of two model list.

### 2.2.3. Models training

The training process of a CNN is an optimization problem in which various functions are applied to measure the distance between the true value vector *y* and the predicted value vector *ŷ*. For classification tasks, Categorical Cross Entropy (*CCE*) is the simplest and most commonly used objective function, shown in Equation (6). For each element in true value *y* and output *ŷ*, *CCE* yield 1 if they are the same, otherwise, it is 0.

$$y = [y_1, y_2 \cdots y_n]^T, \ \hat{y} = [\hat{y}_1, \hat{y}_2 \cdots \hat{y}_n]^T \tag{3}$$

$$equal^*(a, b) = \begin{cases} 1 & a - b = 0 \\ 0 & a - b \neq 0 \end{cases} \tag{4}$$

$$equal([a_1, a_2, \cdots a_n]^T, [b_1, b_2, \cdots b_n]^T) = [equal^*(a_1, b_2), equal^*(a_2, b_2) \cdots equal^*(a_n, b_n)]^T \tag{5}$$

$$CCE = equal(y, \hat{y}) \tag{6}$$

Although *CCE* is widely used, it still has some shortcomings in handling misclassification of farming tasks, because *CCE* leads the model to predict accurately based on a unified rule. In the real framing tasks, the cost of misclassification will depend on the pragmatic situation; i.e., the cost of classifying weed to crop is low while it's bearable to leave a few weeds; however, the cost of classifying crop to weed tends to be high, while it's disadvantageous to remove any part of authentic crops.

In order to lower the risk of killing the crops, new objective functions are used in our training procedure: no miss weed (*NMW*) in Equation (11) and dual metrics (*DM*) in Equation (12). In Equation (7): *w* stands for weed in the data set, *p* for kinds of weed, *q* for kinds of crop in the data set, *u* for the unlabeled object, which does not exist in the original set and has been added for new objective function, and *p, q, r* stand for the volume of their sets. In Equation (8), the result of *contain()* function is a vector if the target vector *tar* contains any elements in template vector *temp*, e.g., the value of *tar* is assumed to be [1,0,2,1] for model's answer, the value of *temp* is assumed to be [0,2] for valid answer, so the result of *contain*() is [False, True, True, False].

$$w = [w_1, w_2 \cdots w_p]^T, \ c = [c_1, c_2 \cdots c_q]^T, \ u = [u_1, u_2 \cdots u_r]^T, \tag{7}$$

$$contain(temp, tar) = \vee_{temp_i}^{temp}[tar \wedge temp_i \cdot 1] \tag{8}$$

The objective function *NMW* is comprised of *CCE*, tolerance of homogeneity $Tol_{homo}$ in Equation (9) and tolerance of unknown $Tol_{unknow}$ in Equation (10). $Tol_{homo}$ yield 1 when predicted value and its true value are in the same group, and $Tol_{unknown}$ yield 1 when predict as unknown. These two objective functions disclose some latent information to models, or some tolerance when models predict with slight errors. Then we apply a logical OR operator between *CCE* and tolerance. In short, *NMW* returns 0 if and only if the object belonging to crop is predicated as a weed.

$$Tol_{homo} = \vee_a^{[w,c]}[contain(a, y) \wedge contain(a, \hat{y})] \tag{9}$$

$$Tol_{unknow} = contain(u, y) \tag{10}$$

$$NMW = Acc \vee (Tol_{homo} \wedge Tol_{unknow}) \tag{11}$$

$$DM = [Acc, NMW] \tag{12}$$

*DM* is a vector of objective functions instead of a single objective function like *CCE* and *NMW*. In model training, if the model's optimizer accepts multiple objective functions, the model will optimize them separately instead of optimizing the corresponding sum. After randomizing weights, we trained these models with these two new objective functions: *NMW* and *DM*.

### 2.3. Ensemble modeling and prediction

The strategy of ensemble modeling is shown in Fig. 2 S4, where models A, B and C have been trained in the same configuration; $P_1$, $P_2$ and $P_3$ as predictions of the models, and the corresponding order is arbitrary; in $Pred_{(m)}$ block, *cate* is category of prediction, like potato, tomato or apple; *type* is the group of this prediction, crop, weed or unknown; and $Act_{(m)}$ shows how the model suggest to deal with these objects. Some key points of the ensemble modeling are explained as follows.
1) If all models reach a consensus, in the sense that they give the same prediction in a specific category, then give the final prediction. 2) If more than half of the models reach a consensus on identifying as a crop, e.g. [$crop_1$,$crop_1$,$weed_1$] of three models, we mark this object as $crop_1$ if there still have a gap between classified $crop_1$ and the total $crop_1$, and 3) Otherwise we merely mark it as unknown.

When dealing with consensus and disagreement, our method is the same as other methods of ensemble models, but we have different treatments in processing crop-like (Case 2 above) and the other predictions, because treating a crop as a weed, which causes crop killing, will reduce the yield of field production. For Case 2, total amount of crops was employed in ensemble strategy since it was easy to obtain in field.

## 3. Result and discussion

### 3.1. Performance metrics

Performance metrics used in this paper include *Accuracy* and *Recall$_{crop}$*, which are defined in Equations (13) and (15).

$$Accuracy = \frac{1}{|w+c|} \cdot |\; equal(y, \hat{y})\;| \tag{13}$$

$$CKR = \frac{1}{|c|} \cdot |\sim equal(y \cdot contain(c, y), \hat{y} \cdot contain(w, \hat{y})\;)| \tag{14}$$

$$Recall_{crop} = 1 - CKR \tag{15}$$

where $y$ denotes the true value, and $\hat{y}$ means the prediction value. Besides, function *equal*, function *contain*, weed data $w$, and crop data $c$ are defined in Equations (5), (7) and (8) earlier.

In crop and weed classification, *Accuracy* is the number of correctly predicted plants out of all the plants. For *CKR*, its numerator is the number of crops identified as weed, and its denominator is the number of all crops. *Recall$_{crop}$* is the detected crops out of all crops, which equals to $1 - CKR$. Note that identifying a crop to other kind of crop is a valid detection in the context of *Recall$_{crop}$*.

### 3.2. Models from searching

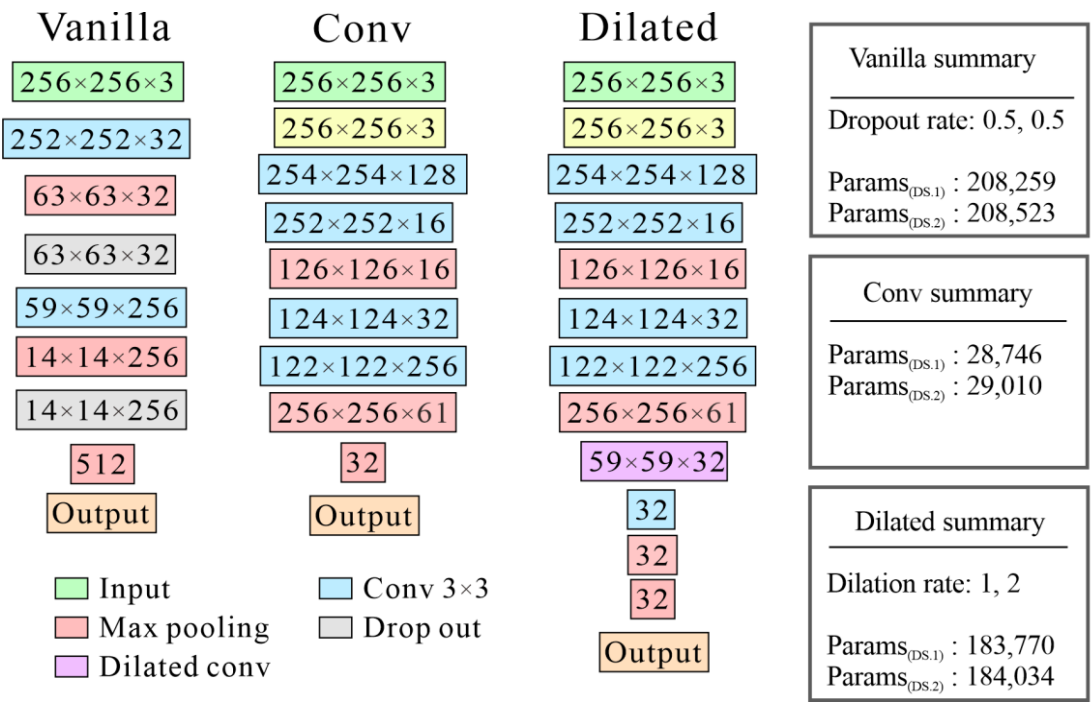

**Figure 4.** Configuration of most appropriate models by Algorithm 2.

For brevity of description, we name the generated models by their certain characteristics and structures. Vanilla neural network, dubbed Vanilla, has similar structure as an earlier neural network named AlexNet [13]. Convolution neural network, dubbed Conv, has a similar structure as a famous convolution neural networks named VGG [14]. Dilated convolution neural network, dubbed Dilated, has a similar structure as Conv but a new type of dilated convolution block [15,16].

To evaluate these three generated models by AutoML, five state of the art CNN models are used as baseline models for comparison: DenseNet201 [17], InceptionV3 [18], VGG19 [14], Xception [19] and Res-Net152v2 [20]. These five models have achieved the top ranking in the ImageNet competition [21]. Their number of parameters range from twenty million to one hundred million, at least 100 times of our generated model parameters. All the baseline models are denoted with * in the last five rows of Table 1.

Because crops are the goal of agricultural production, the mis-identification of crops as weeds can be costlier than other cases. *Accuracy* measures only overall performance on crop and weed, so we use $Recall_{crop}$ to measure the effect on crop classification. For example, suppose a dataset has two classes dubbed as CROP and WEED containing 7 weeds and 3 crops. If a crop and a weed were identified incorrectly, then we would get 80% *Accuracy* and 60% $Recall_{crop}$ according to (13) and (15). The training time of the baseline model and the search time of the generated model are also considered. Details of the evaluation are shown in Table 1.

**Table 1.** Models comparison.

| Model | DS.1 | | | DS.2 | | |
|---|---|---|---|---|---|---|
| | Accuracy | Recall$_{crop}$ | Time | Accuracy | Recall$_{crop}$ | Time |
| Vanilla | 98.06% | 93.36% | 12.4h | 99.24% | 97.76% | 15.5h |
| Dilated | 98.34% | 94.24% | 9.5h | 99.53% | 98.64% | 11.3h |
| Conv | 97.51% | 93.72% | 12.4h | 99.06% | 96.88% | 14.2h |
| *DenseNet201 [17] | 98.26% | 93.88% | 0.6h | 98.95% | 96.93% | 0.5h |
| *InceptionV3 [18] | 98.96% | 94.93% | 0.6h | 99.51% | 97.49% | 0.6h |
| *VGG19 [14] | 98.58% | 94.15% | 0.8h | 99.20% | 96.59% | 0.8h |
| *Xception [19] | 97.16% | 92.65% | 0.7h | 99.49% | 97.77% | 0.6h |
| *ResNet152v2 [20] | 99.52% | 98.44% | 1.0h | 99.58% | 98.72% | 1.0h |

The above models achieve high *Accuracy* and $Recall_{crop}$ at most cases. For the training time on GTX 2080Ti GPU, it took more than 10 hours to generate the model, while the baseline model only took less than 1 hour. Such time consumption is acceptable if no human intervention is required.

However, the $Recall_{crop}$ score is always lower than the *Accuracy* score, which may be caused by prediction bias in imbalanced data sets. Low $Recall_{crop}$ is an unpromising result for crop and weed classification. In addition, *CCE* is used to determine the specific type of target, which assumes that all misclassification errors made by a model are equal. Thus, we tried to use sampling approaches to treat the imbalanced data set, and new objective functions to fix the training problem regarding *CCE*.

According to *Accuracy* and $Recall_{crop}$ values in Table 1, two scatter diagrams are plotted to further analyze the performances of these models in Fig. 5. We can see that ResNet achieves the best performance, hence we choose ResNet as our baseline models in the subsequent experiments. Limited to parameter size, the performances of the generated models are similar to most baseline models.

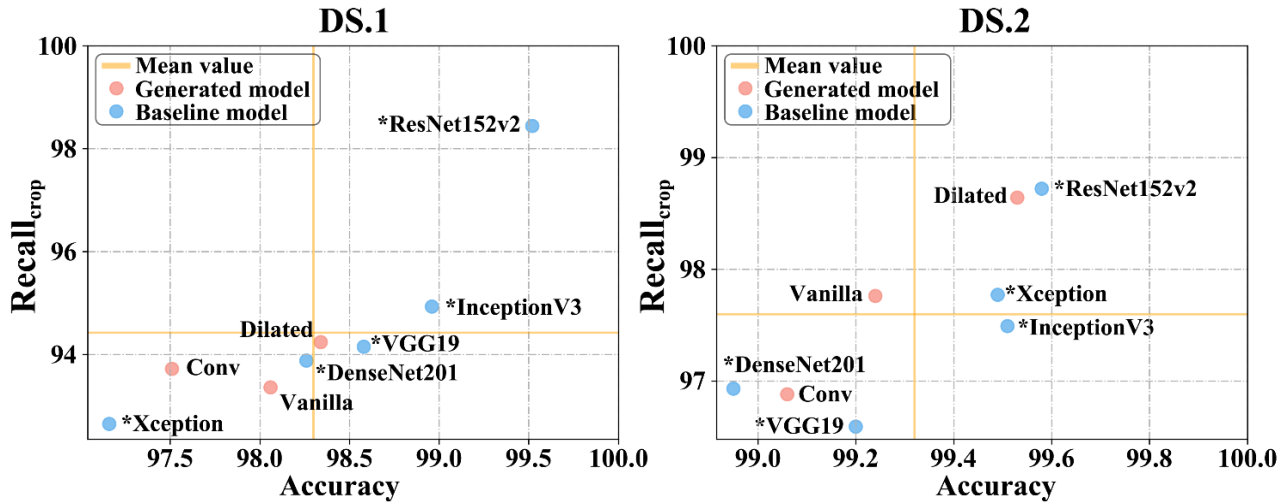

**Figure 5.** Accuracy and Recall_crop of the models.

### 3.3. Model training

The training curves of the models and the corresponding final scores are shown in Fig. 6. For each data set, *DS.1* and *DS.2*, model training is carried out based on sampled data set and complete data set. For each data usage method, the models are trained considering three objective functions i.e., *CCE*, *NMW* and *DM*. As *CCE* and *NMW* are combined in *DM*, it has two separate plots which are *DM-CCE* and *DM-NMW*.

In general, the curves become stable after about 60 epochs, reaching high scores of more than 96% after 64 epochs. The curves of ResNet have more fluctuations than others, which might be due to the corresponding deep and special structures. The final scores of *NMW* are slightly higher than scores of *CCE*, while *DM-CCE* scores slightly higher than *CCE* scores. Based on the training curves and scores, the models are sophisticated enough to be applied for the next task. So, we use the combined model to determine the appropriate configuration, in other words, the one with high *Accuracy* and low *CKR*.

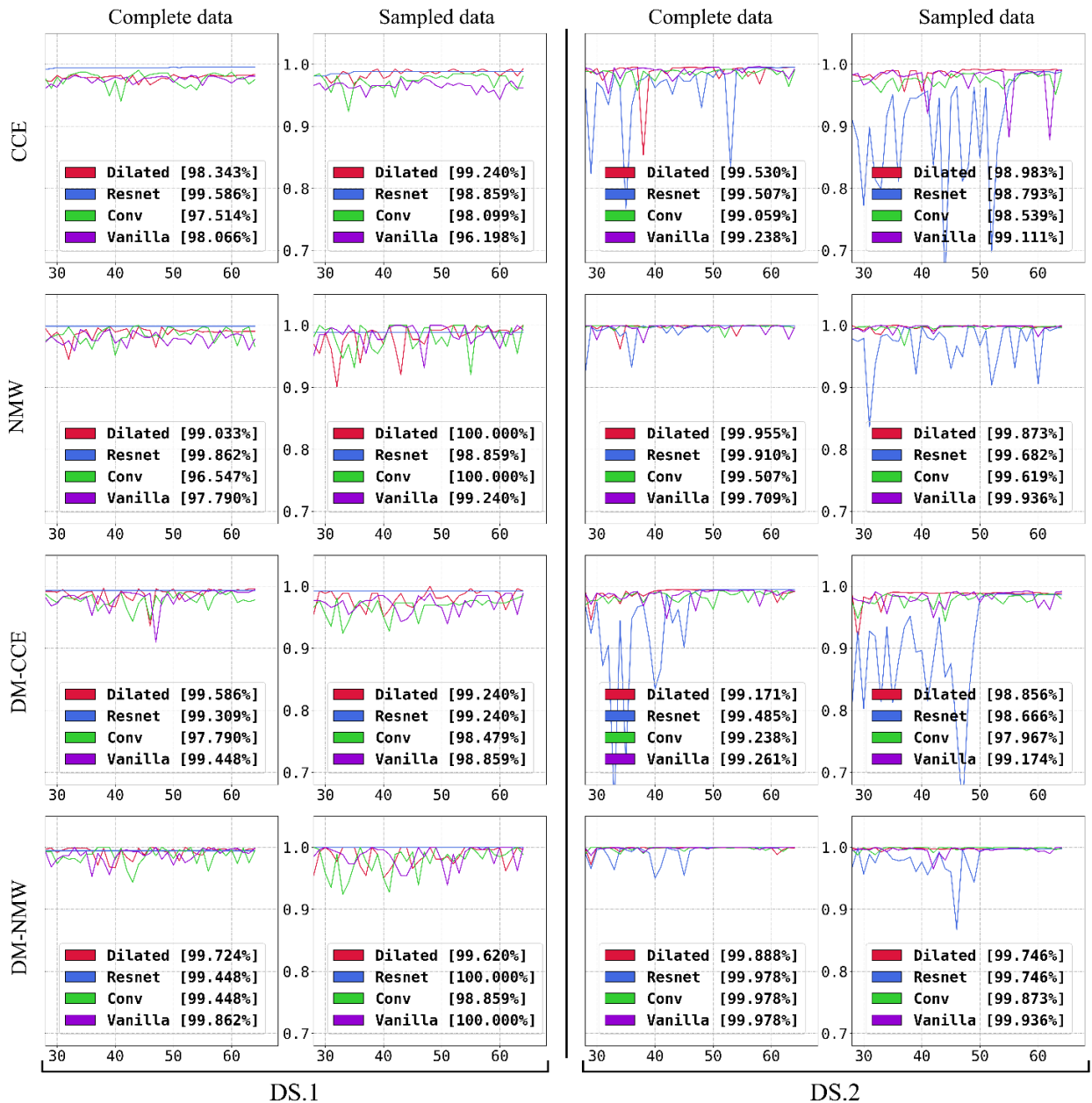

**Figure 6.** Models training.

## 3.4. *Ensemble modeling and predicting*

The ensemble strategy shown in Section 2.3 is used to build the ensemble model from three generated models. Since there are two data usages (CD and SD) and three objective functions (*CCE*, *NMW* and *DM*), we have six configurations on each data set, as shown in the first column of Table 2. Evaluations of ensemble models and the corresponding sub-models on the test data sets are shown in Table2 and Fig. 6.

**Table 2.** Accuracy of models.

| Experiments | DS.1 | | DS.1 | |
|---|---|---|---|---|
| | **The best** | **Ensemble** | **The best** | **Ensemble** |
| CD-CCE | 99.58% | 99.27% | 99.53% | 99.02% |
| CD-NMW | 99.03% | 99.76% | 99.41% | 99.64% |
| CD-DM | 99.58% | 99.76% | 99.48% | 99.84% |
| SD-CCE | 99.24% | 99.52% | 99.11% | 99.92% |
| SD-NMW | 99.27% | 99.76% | 99.56% | 99.88% |
| SD-DM | 99.24% | 99.76% | 99.87% | 99.96% |

In the evaluation, the scores of the sub-models are no longer displayed separately except for the highest and lowest ones in the sub-models. In Table 2, 'The Best' indicated the highest *Accuracy* among the sub-models, which are slightly better than the ensemble model in most cases.

Besides achieving high *Accuracy*, analyzing the misclassification to reduce *CKR* is the main goal of this project, hence 0% CKR means that no crop is misclassified as weed, so no crop will be wrongly killed. To measure the misclassification more accurately, we divide them into four categories as follows:

1. Moderate errors: they occur when the model classifies an object as unknown, which occur in new objective functions.

2. Minor errors: they occur when the model's prediction is inconsistent with the corresponding label, but both are in the same type, for instance the prediction is weed$_1$, but the label is weed$_2$;

3. Considerable errors: they occur when the models predict a weed as a crop;

4. Dangerous errors: they occur when the models predict a crop as a weed.

These errors are ranked based on their outcomes in practice. For moderate errors, we have a chance to rectify them manually. While minor errors do little (or nil) harm to the field, considerable errors leave a weed in field and dangerous ones kill a crop mistakenly.

By dividing the misclassification, we find that *CKR* only depends on the amount of dangerous errors. For a specific crop named A, the misclassification types would include classifying A as a weed (dangerous error), classifying A as another crop (minor error), and classifying A as an unknown object (moderate error). Thence, the total *CKR* will decrease if the dangerous errors can be reduced.

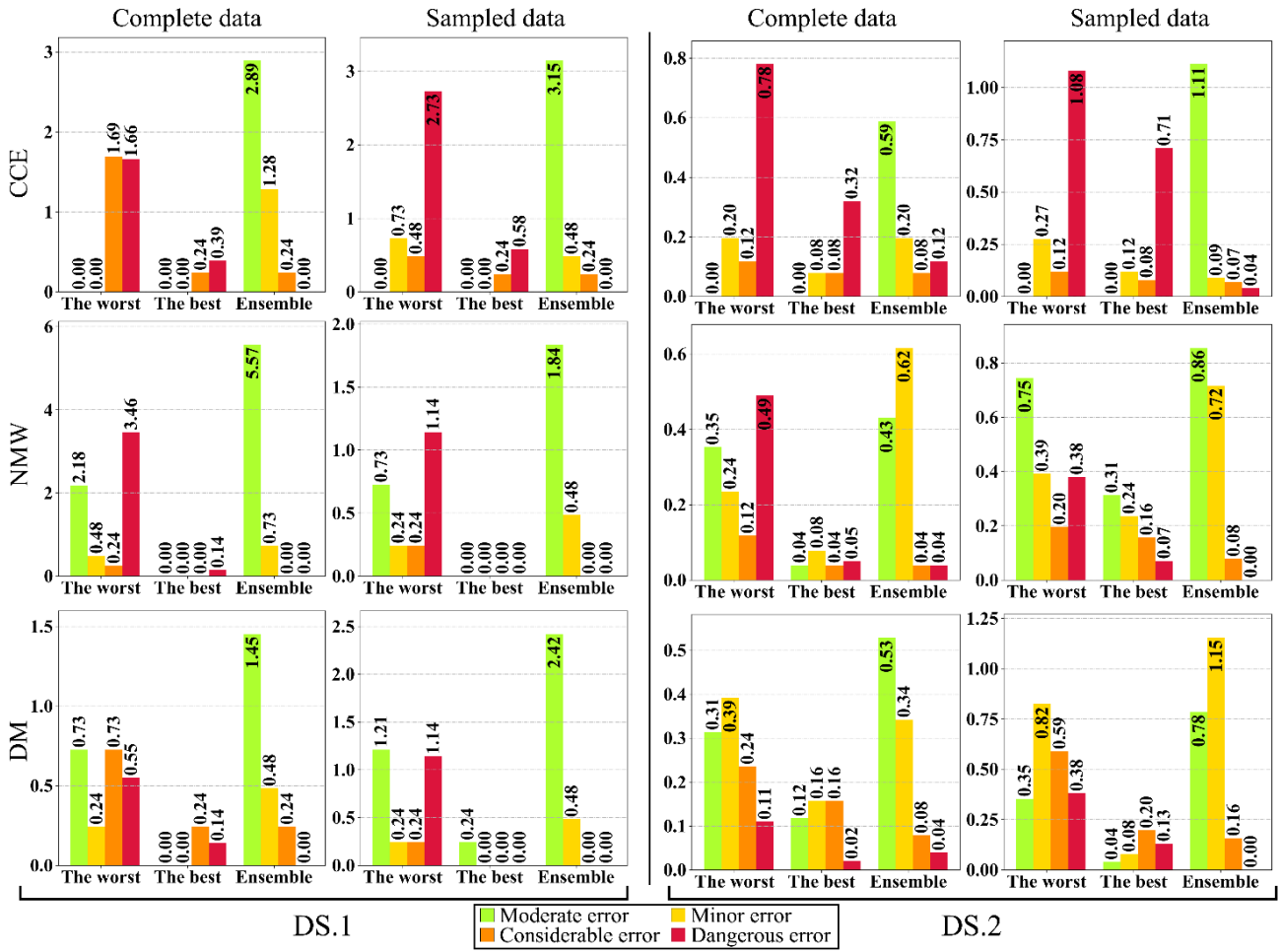

**Figure 7.** Errors percentages of models.

The percentages of four errors are shown in Fig. 7. In the analysis of errors' percentage, the worst means the highest scores of four errors, in other words, the four scores may come from different sub-models. Accordingly, the best means the lowest scores herein.

For sub-models, the models with *CCE* reach the highest dangerous error rates, and the models with *DM* reach the lowest dangerous error rates. For data usage, sampling approaches did not really achieve an obvious reduction in error rates. In most cases, the ensemble model reaches the lowest or even 0.00% dangerous error rate, and the highest moderate error rate. In summary, our proposed ensemble strategy and the corresponding objective function can reduce the *CKR* by lowering the rate of dangerous errors.

## 4. Conclusions

This work proposes a methodology of crop and weed classification based on AutoML and ensemble modeling. AutoML-based algorithm helps us to automatically choose the CNN models among two data sets. Models with different data usages and different objective function are used to build ensemble models. Overall, the ensemble model with objective function *DM* achieves highest *Accuracy* and lowest *CKR*. Thus, we hypothesize that by applying this method, we can effectively move towards desired outcomes in precision farming.

While we made contributions as mentioned above, due to the environmental constraints, our method was only evaluated in green house, and performance in other environments still needs to be tested. We hope to tackle the limitation in near future experiments with input of more generic data sets.

## Acknowledgments

This work was partially supported by National Key R&D Program of China under Grant No. 2020YFC0832500, Ministry of Education - China Mobile Research Foundation under Grant No. MCM20170206, The Fundamental Research Funds for the Central Universities under Grant No. lzujbky-2021-sp47, lzujbky-2020-sp02, lzujbky-2019-kb51 and lzujbky-2018-k12, National Natural Science Foundation of China under Grant No. 61402210, State Grid Corporation of China Science and Technology Project under Grant No. SGGSKY00WYJS2000062, Science and Technology Plan of Qinghai Province under Grant No.2020-GX-164, Google Research Awards and Google Faculty Award. We also gratefully acknowledge the support of NVIDIA Corporation with the donation of the Jetson TX1 used for this research.

## Conflict of interest

A request to access data can be directed to authors. The research performed in this work is the sole work of the named authors. The ideas presented in this article do not pose any risks to individuals or institution. We declare that we do not have any conflicts of interest regarding the study.

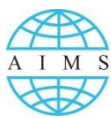 AIMS Press